\documentclass{article}

\usepackage[final,nonatbib]{neurips_2022}

\usepackage[utf8]{inputenc} 
\usepackage[T1]{fontenc}    
\usepackage{url}            
\usepackage{booktabs}       
\usepackage{amsfonts}       
\usepackage{nicefrac}       
\usepackage{microtype}      
\usepackage{xcolor}         
\usepackage{amsmath}
\usepackage[toc,page]{appendix}
\usepackage{graphicx,subcaption}
\usepackage{caption}

\title{Conditional Neural Processes for Molecules}

\author{%
  Miguel Garcia-Ortegon \\
  DPMMS\\
  University of Cambridge\\
  Wilberforce Rd, Cambridge, UK \\
  \texttt{mg770@cam.ac.uk} \\
  \And
  Andreas Bender \\
  Dept. of Chemistry\\
  University of Cambridge\\
  Lensfield Rd, Cambridge, UK \\
  \texttt{ab454@cam.ac.uk} \\
  \And
  Sergio Bacallado \\
  DPMMS\\
  University of Cambridge\\
  Wilberforce Rd, Cambridge, UK \\
  \texttt{sb2116@cam.ac.uk} \\
}

\begin{document}

\maketitle

\begin{abstract}
    Neural processes (NPs) are models for transfer learning with properties reminiscent of Gaussian Processes (GPs). They are adept at modelling data consisting of few observations of many related functions on the same input space and are trained by minimizing a variational objective, which is computationally much less expensive than the Bayesian updating required by GPs. So far, most studies of NPs have focused on low-dimensional datasets which are not representative of realistic transfer learning tasks. Drug discovery is one application area that is characterized by datasets consisting of many chemical properties or functions which are sparsely observed, yet depend on shared features or representations of the molecular inputs. This paper applies the conditional neural process (CNP) to \textsc{dockstring}, a dataset of docking scores for benchmarking ML models. CNPs show competitive performance in few-shot learning tasks relative to supervised learning baselines common in chemoinformatics, as well as an alternative model for transfer learning based on pre-training and refining neural network regressors. We present a Bayesian optimization experiment which showcases the probabilistic nature of CNPs and discuss shortcomings of the model in uncertainty quantification. 
\end{abstract}

\vspace{-0.125cm}

\section{Introduction}

\subsection{Learning from sparse chemical datasets}

Recent years have seen an explosion of novel machine learning (ML) methods for molecular tasks, often relying on large neural networks that require vast amounts of labelled data. The development of these models has been fueled by an expectation that ML will greatly accelerate drug discovery~\cite{bender, chapter}. Unfortunately, their real-world applicability is hindered by the sparsity of chemical datasets, which comprise many molecular functions with a few observations each. It is estimated that in-house datasets in the pharmaceutical industry are less than 1\% complete, whereas ChEMBL is only 0.05\% complete~\cite{optibrium}. In order to take advantage real-world chemical datasets, we require models that are able to transfer information across separate functions, even if annotated on non-overlapping molecules, and can make predictions on new functions with very few observed labels. This setting, known as meta-learning, could be used to frame and make an impact on problems in many areas of computer-aided drug design, including virtual screening, data imputation, quantitative structure-activity relationships (QSAR), Bayesian optimization or bioactivity fingerprinting, among others.

Neural processes are a novel family of models that show promise in meta-learning but have so far only been tested on toy low-dimensional datasets. In this paper, we evaluate the performance of the CNP in several molecular tasks using high-dimensional molecular representations.

\subsection{Conditional Neural Processes (CNPs)}

Consider a dataset consisting of observations of real-valued functions $f_1,\dots,f_n$ on the same input space $\mathcal X$. Each function $f_i$ is observed at a set of $o_i$ input points $x_O^i\in \mathcal X^{o_i}$; we define $y_O^i = (f(x^i_{O,1}),\dots,f(x^i_{O,o_i}))$. Let $f$ be a \emph{test} function, $(x_C,y_C)$ be a vector of $c$ \emph{context} points and the values of $f$ on these inputs, and $(x_T,y_T)$ be a vector of $t$ \emph{target} points and the values of $f$ on these inputs. A neural process (NP) ~\cite{cnp, lnp} aims to describe the predictive distribution $p(y_T \mid x_T, x_C, y_C)$. This is done by mapping $(x_C,y_C,x_T)$ through a parametric function, which is trained on the data $(x_O^i,y_O^i)_{i=1,\dots,n}$. In particular, we model the predictive distribution with a product measure:
$$q_\theta(y_T \mid x_T, x_C, y_C) = \prod_{j=1}^t \mathcal N(y_{T,j}; \mu_\theta(x_{T,j}),\sigma_\theta^2(x_{T,j})).$$
The mean and variance, $\mu_\theta(x)$ and $\sigma_\theta^2(x)$, of the predictive distribution at target input $x$ are obtained through the following mapping:
\begin{align*}
    r_j &= h_\theta(x_{C,j},y_{C,j}) \qquad \text{ for }j=1,\dots,c &  \text{(encoding)} \\
    r &= r_1 \oplus \cdots \oplus r_c &  \text{(aggregation)} \\
    (\mu_\theta(x),\sigma_\theta^2(x)) &= g_\theta(x,r) \qquad ~~~~~~~~~~~\text{ for all } x\in\mathcal X &  \text{(decoding)}
\end{align*}


where $h_\theta$ and $g_\theta$ are neural networks, $\oplus$ is a commutative operation and $r$ is a global representation for the entire context data.This architecture ensures that the predictive distribution is invariant to permutations of the context and target points, respectively. The parameters $\theta$ of the encoder and decoder are trained by backpropagation using the data $(x_O^i,y_O^i)_{i=1,\dots,n}$. Conditional NPs (CNPs)~\cite{cnp} minimise a particularly simple objective:
$$
- \mathbb{E} \left[ \frac{1}{n} \sum_{i=1}^n \log q_\theta(y^i_T \mid x_T^i, x_C^i, y_C^i) \right],
$$
where the expectation is taken with respect to a random partition of the observations $(x_O^i,y_O^i)$ for function $f_i$ into a set of context points $(x_C^i,y_C^i)$ and target points $(x_T^i,y_T^i)$. This objective function does not explicitly regularize the predictive distribution $q_\theta$, so when the model is overparametrized, the objective can diverge and the variance parameters $\sigma^2(\cdot,r)$ can underestimate uncertainty. Latent NPs (LNPs)~\cite{lnp} avoid this problem by maximizing an approximate Evidence Lower Bound, derived through a more conventional variational inference approach. 

So far NP models have been evaluated on low-dimensional settings, where they excel at few-shot learning. Here, we will analyze their performance on molecules represented by high-dimensional chemical fingerprints.

\subsection{The \textsc{DOCKSTRING} dataset}

The \textsc{dockstring} dataset~\cite{dockstring} is a molecular dataset for benchmarking of ML models. It comprises more than 15 million docking scores for 58 protein targets and 260k molecules. Targets were chosen to be medically relevant and represent a variety of protein families, and molecules were curated from PubChem and ChEMBL to be representative of chemical series in drug discovery projects. 

Each molecule in the \textsc{dockstring} dataset is annotated with all 58 protein target scores, which makes it especially suitable to design benchmark tasks in transfer learning. In this paper, we sample a small subset of it to evaluate regression and transfer learning by CNPs in the low-data regime.


\section{Methods}

\subsection{Dataset and split}

\begin{figure}[h]
    \centering
    \vspace{-0.1cm}
    \includegraphics[trim=15 14 0 0, clip, width=0.55\textwidth]{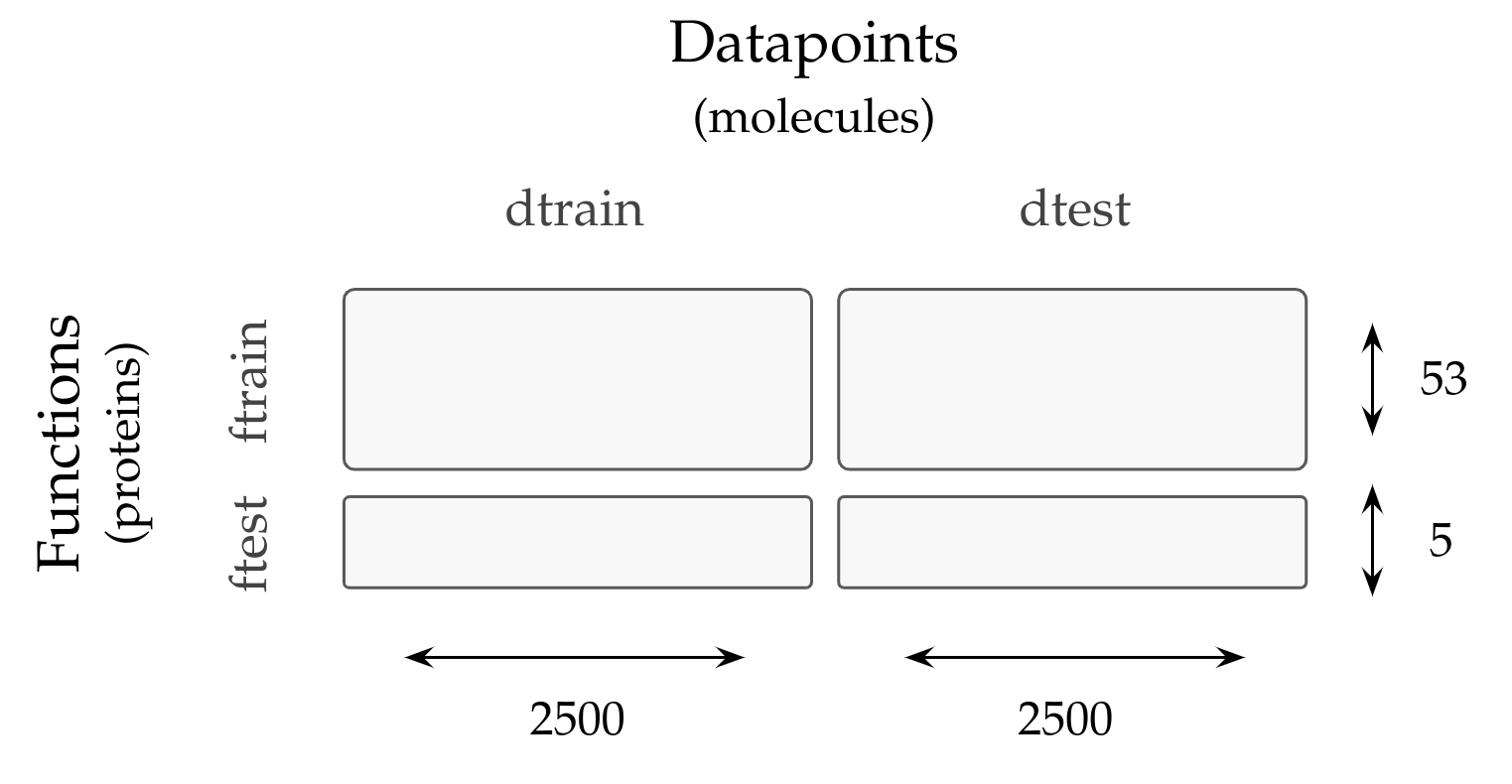}
    \caption{Split across the function and datapoint dimensions of the \textsc{dockstring} sample. \textit{dtrain} was sampled from the \textsc{dockstring} training set and \textit{dtest} was sampled from the \textsc{dockstring} test set, which were split by cluster similarity \cite{dockstring}. \textit{ftrain} and \textit{ftest} were derived from the \textsc{dockstring} regression tassk}
    \label{fig:split}
\end{figure}

NPs are able to learn across different datapoints within the same function, and across different functions within the same input space. Therefore, the dataset was split across both the datapoint dimension and the function dimension. We refer to these splits as \textit{dtrain}, \textit{dtest} and \textit{ftrain}, \textit{ftest} respectively (Figure~\ref{fig:split}).

To emulate learning in a low-data regime, we took a small sample of the train and test sets defined in the \textsc{dockstring} package. The \textit{dtrain} set consisted of 2500 molecules from the original train set, and the \textit{dtest} set consisted of 2500 molecules from the original test set. \textsc{dockstring} original sets were split by clusters, which prevented data leakage from chemical analogues in \textit{dtrain} and \textit{dtest}. Our function split was derived from the \textsc{dockstring} regression task, using the 5 task targets (ESR2, KIT, PARP1, PGR, F2) as \textit{ftest} and the other 53 targets as \textit{ftrain}.

\subsection{CNP and benchmark models}

A simple CNP was implemented with 3 linear layers in the encoder network, a mean aggregator function and 3 linear layers in the decoder network. We include four benchmarks commonly used in ML for chemoinformatics: a feed-forward neural network with the same number of layers as the CNP (NN),  k-nearest neighbours with $k=5$ (KNN) and $k=1$ (FSS, known as fingerprint similarity search in chemoinformatics~\cite{fss}),  and a random forest regressor with 200 estimators (RF). As these benchmarks are only trained on \emph{ftest} functions, we include a further benchmark for transfer learning (fine-tuned NN). This consists of pre-training the previous NN with 53 outputs on \emph{ftrain} observations, and fine-tuning the model on each \emph{ftest} function. The input to all models were Morgan molecular fingerprints of radius 3 and length 1024. The CNP and NN models were implemented in Pytorch~\cite{pytorch}, and the rest were implemented in scikit-learn~\cite{scikit-learn}. 


\section{Experiments}

\subsection{Probabilistic regression and calibration}

The CNP is an overparameterized neural model that outputs a predictive distribution. Similarly to neural networks trained by maximum likelihood, the CNP is trained by maximizing the conditional probability of the target points given the context points. However, unlike most neural models, the CNP performs uncertainty quantification. Since the conditional probability of the target points could be made arbitrarily large by making the predicted variance smaller, the CNP is at risk of overfitting and producing unreliable uncertainty estimates.

\begin{figure}[h!]
\vspace{0.2cm}
\captionsetup[subfloat]{labelformat=empty}
\centering
\begin{minipage}{1\linewidth}
\begin{minipage}{.5\linewidth}
\centering
\subfloat[]{\label{fig:regression_calibration_a_improved}\includegraphics[scale=.5]{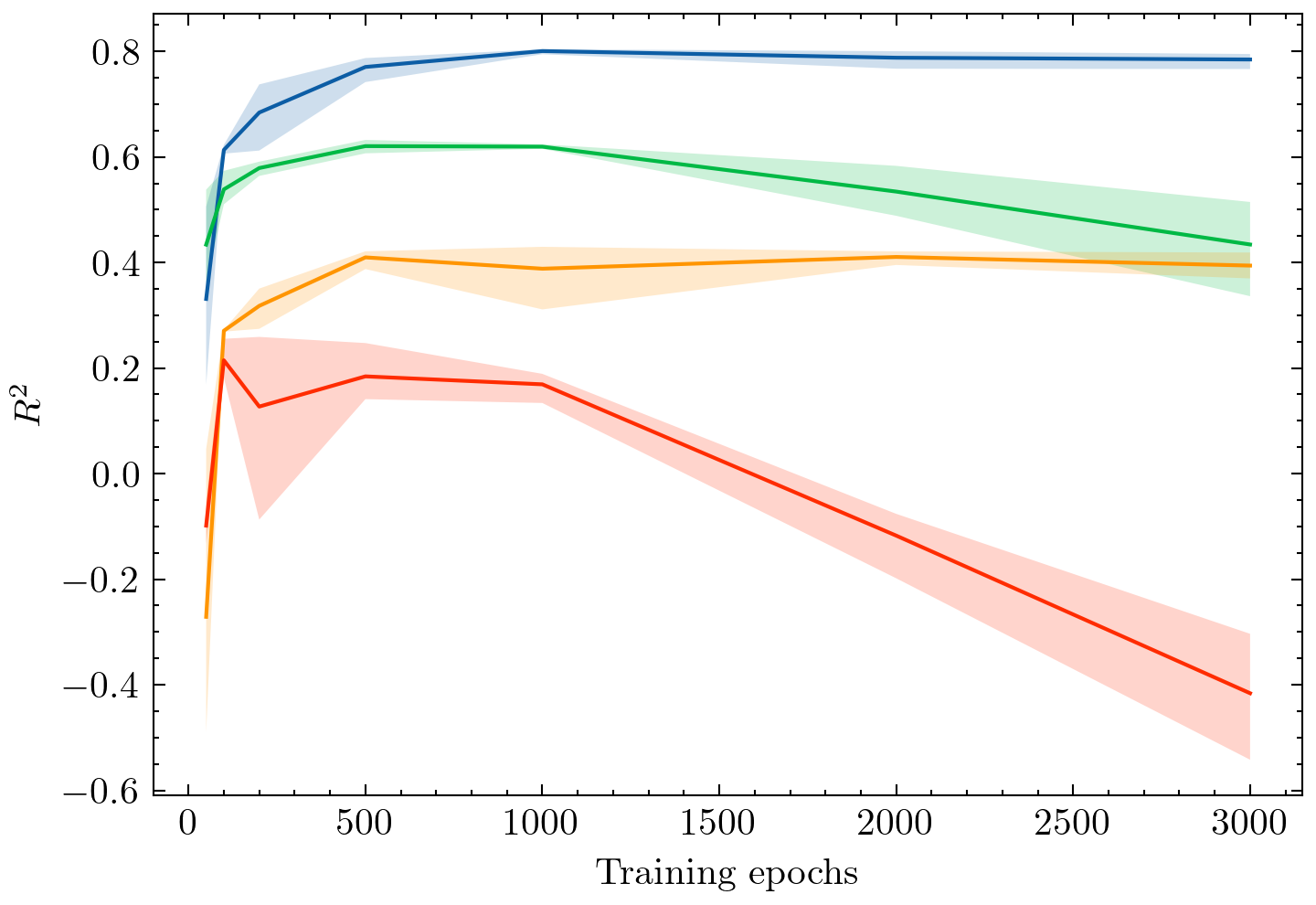}}
\end{minipage}
\begin{minipage}{.5\linewidth}
\centering
\subfloat[]{\label{fig:regression_calibration_b_improved}\includegraphics[scale=.5]{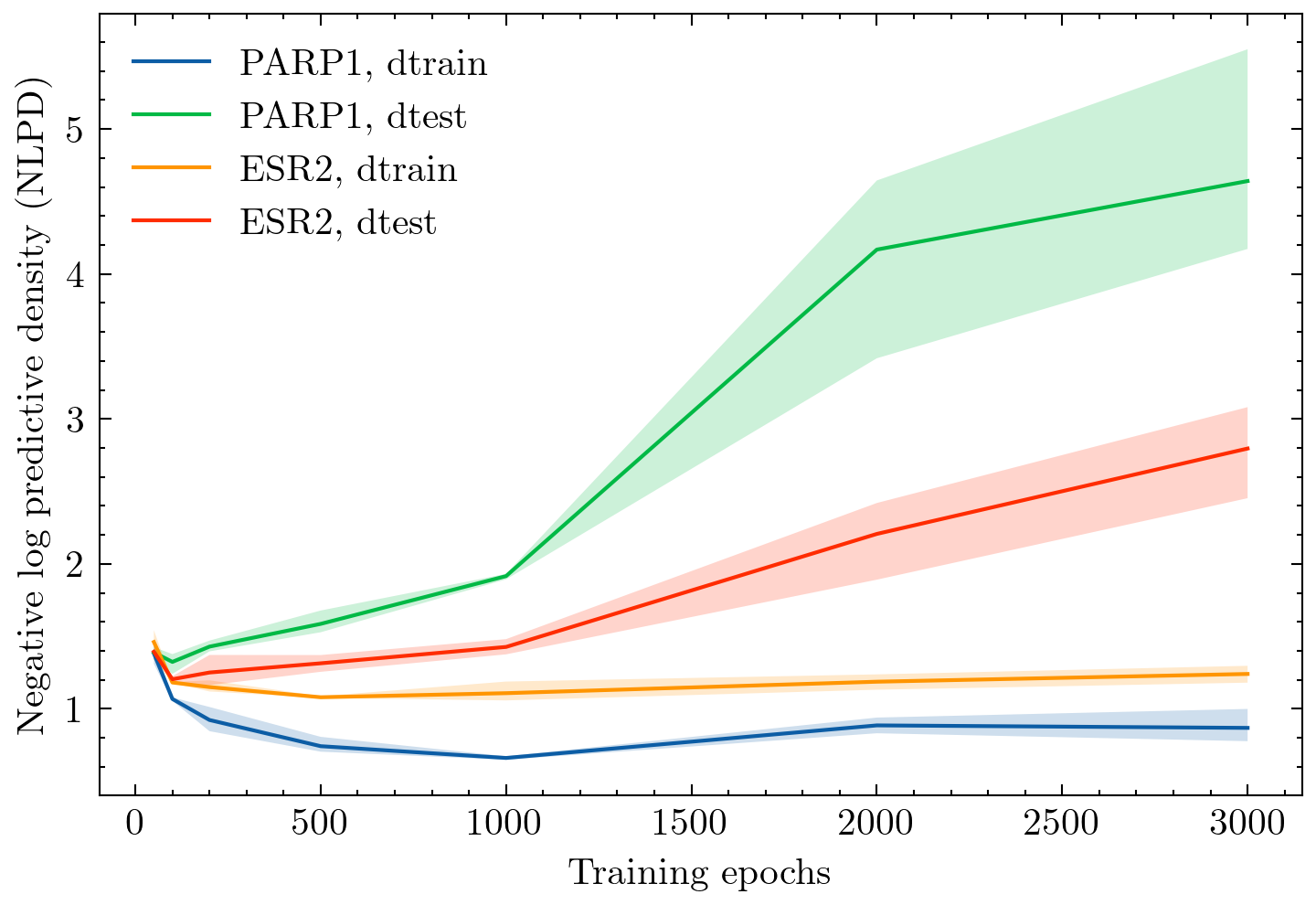}}
\end{minipage}
\end{minipage}
\vspace{-0.2cm}
\caption{CNP performance on two \textit{ftest} functions (an easy one and a challenging one) as the training time increases. Lines indicate the mean across three random initializations, and shaded regions indicate the maximum and minimum at each training length. Regression performance on Vina docking scores was comparable to that previously reported~\cite{dockstring}.}
\label{fig:regression_calibration_improved}
\end{figure}

To evaluate this phenomenon, we analyzed the regression performance and the negative log predictive density (NLPD) of CNP predictions as the number of training epochs increased (Figure~\ref{fig:regression_calibration_improved}, Appendix~\ref{app:regression_calibration}). Figure~\ref{fig:regression_calibration_improved} shows an example \textit{ftest} protein target that is comparatively easy to predict (PARP1) and an example \textit{ftest} protein target that is challenging (ESR2). We observed that prediction performance on training datapoints improved monotonically or stayed on a similar range as the training time increased, whereas performance on test datapoints degraded after a number of epochs. Similarly, uncertainty estimates remained acceptable for the training datapoints, but worsened dramatically for test datapoints. This discrepancy between training and test datapoints was also observed in \textit{ftrain} functions (Appendix~\ref{app:regression_calibration}) and could be explained by overfitting. Based on these results, we selected the CNP trained for 500 epochs for subsequent experiments.


We hypothesize that the LNP, whose ELBO-like objective includes a KL regularization term~\cite{lnp}, may be more robust to overfitting and degradation of its uncertainty estimates. Analysis of the LNP is left for future work.

\subsection{Few-shot learning}

\begin{figure}[h!]
\captionsetup[subfloat]{labelformat=empty}
\centering
\begin{minipage}{0.9\linewidth}
\begin{minipage}{.5\linewidth}
\centering
\subfloat[]{\label{fig:low_data_a_improved}\includegraphics[scale=.5]{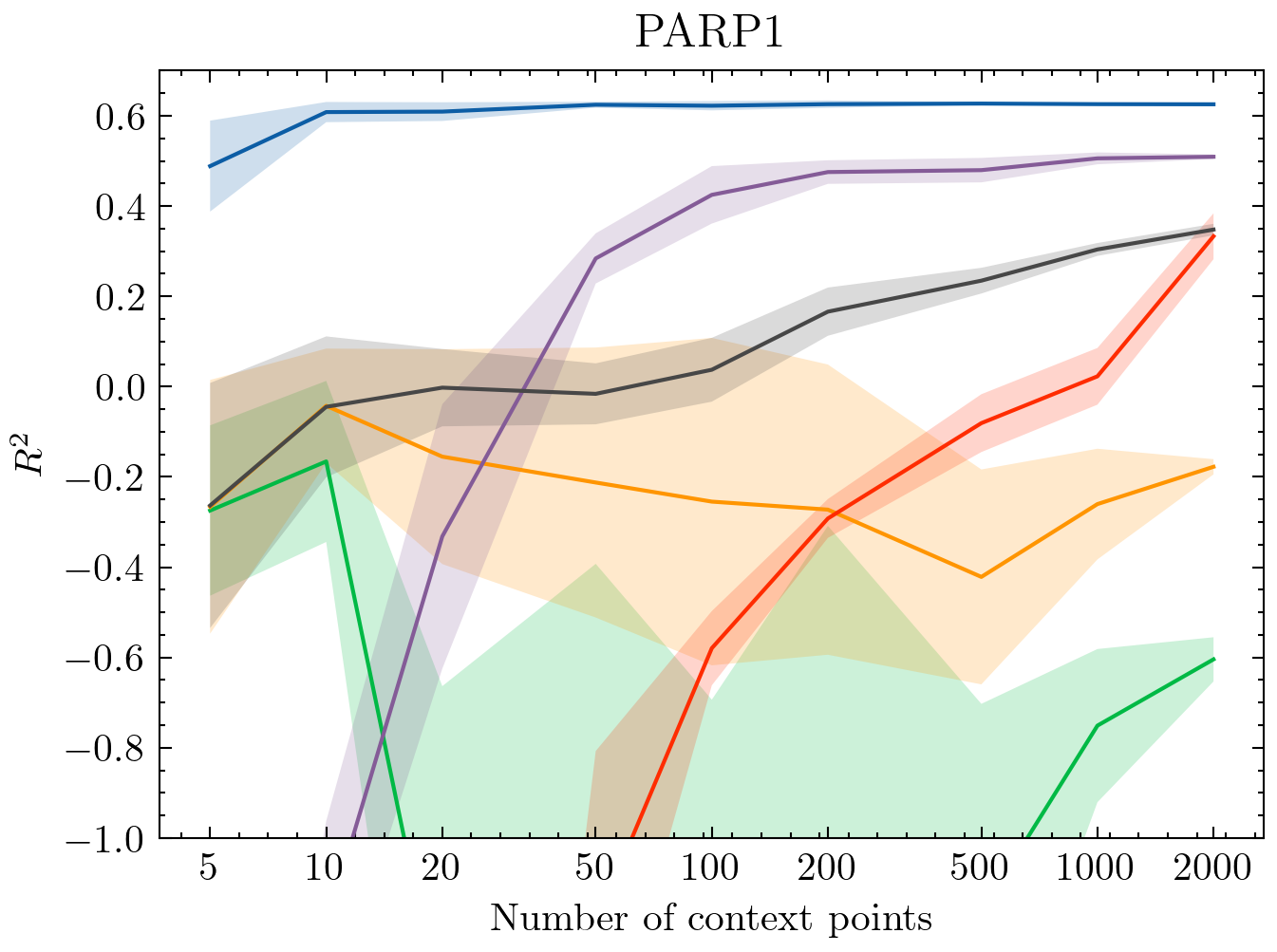}}
\end{minipage}%
\begin{minipage}{.5\linewidth}
\centering
\subfloat[]{\label{fig:low_data_b_improved}\includegraphics[scale=.5]{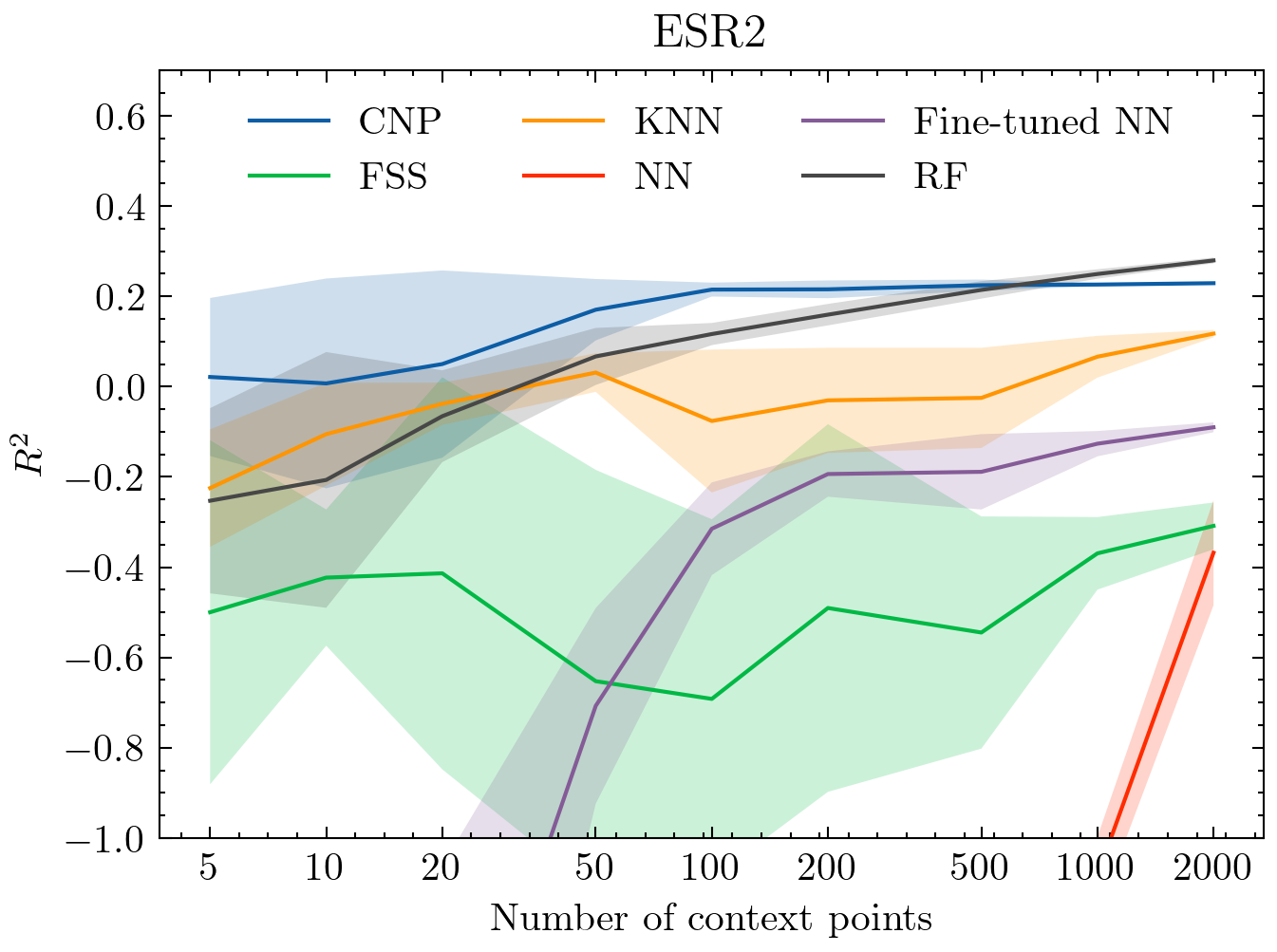}}
\end{minipage}
\end{minipage}
\vspace{-0.25cm}
\caption{Few-shot performance of CNP and benchmark models on two \textit{ftest} functions, an easy one (PARP1) and a challenging one (ESR2). The x-axis shows the number of points in \textit{ftest, dtrain} used as context by the CNP or as training points by the benchmark models. Lines indicate the mean and shaded regions indicate standard deviation across different training runs (in the benchmark models) or across different random selections of context points (in the CNP).
}
\label{fig:low_data_improved}
\end{figure}

We evaluated the performance of the CNP in few-shot learning and compared it against benchmarks popular in chemoinformatics (Figure~\ref{fig:low_data_improved}, Appendix~\ref{app:low_data}). The CNP was trained on \textit{ftrain, dtrain}, used context points in \textit{ftest, dtrain} and was tested on the target points \textit{ftest, dtest}. Other models were trained on \textit{ftest, dtrain} and tested on \textit{ftest, dtest}. In spite of the CNP not seeing the functions in \textit{ftest} during training, it outperformed all other models in the low-data regime. This was the case even for the transfer learning benchmark, fine-tuned NN, which was pre-trained on \textit{ftrain, dtrain} and fine-tuned on \textit{ftest, dtrain}.


\subsection{Generalization to unseen functions}

In previous sections, we analyzed the ability of the CNP to generalize to unseen functions in \textit{ftest}. However, \textit{ftrain} and \textit{ftest} were all part of the same class of Vina docking scores. The question remained whether similar generalization would be observed for molecular functions from very different classes. To investigate this, we created a new type of score that linearly combines docking scores with the quantitative estimate of drug-likeness (QED) (Appendix~\ref{app:qed_regression})~\cite{qed}. We trained CNP models either on plain scores or on plain scores and QED-modified scores, and tested either on plain scores or on QED-modified scores (Table~\ref{tab:qed_regression}). We observed that the CNP was unable to generalize to functions of different classes, but performance could be easily recovered by including functions from those classes in the training set.


    
    

\begin{table}[h!]
  \caption{$R^2$ of CNP trained (\textbf{row}) and tested (\textbf{columns}) on plain and QED-modified scores from PARP1, KIT and F2.}
  \label{tab:qed_regression}
  \vspace{0.3cm}
\centering
\begin{tabular}{ccc}
\toprule
{} &        Plain scores &           QED-modified scores \\
\midrule
Plain scores &  0.57 ± 0.08 &  -6.42 ± 3.14 \\
Plain and QED-modified scores   &  0.54 ± 0.08 &   0.34 ± 0.03 \\
\bottomrule
\end{tabular}
\end{table}

\subsection{Bayesian optimization with CNPs}

\begin{figure}[h]
\captionsetup[subfloat]{labelformat=empty}
\centering
\begin{minipage}{1\linewidth}
\begin{minipage}{.45\linewidth}
\centering
\subfloat[]{\includegraphics[scale=.7]{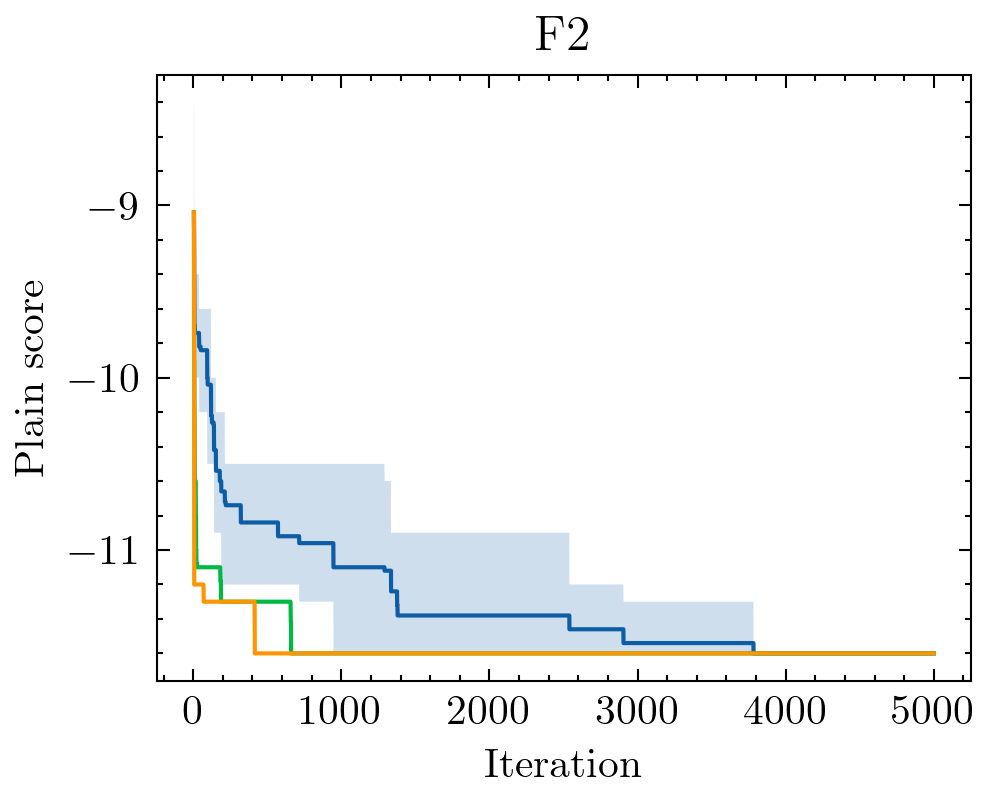}}
\end{minipage}%
\begin{minipage}{.45\linewidth}
\centering
\subfloat[]{\includegraphics[scale=.7]{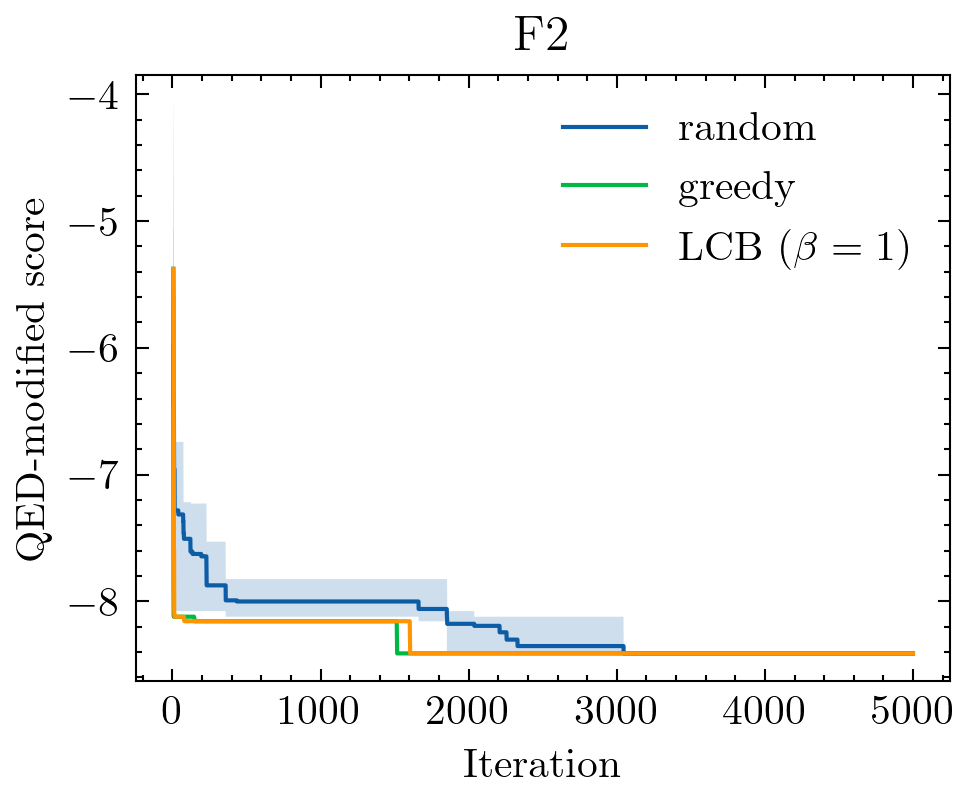}}
\end{minipage}
\end{minipage}
\caption{Bayesian optimization of plain and QED-modified scores of F2, a protein target in \textit{ftest}.}
\label{fig:bo}
\end{figure}

Finally, we evaluated whether the CNP predictive distribution is useful for Bayesian optimization (BO). Plain scores or QED-modifed scores were minimized starting from five context molecules in \textit{ftest, dtrain} $\cup$ \textit{ftest, dtest}, selecting one molecule per iteration for 4995 iterations. Three acquisition strategies to select the next optimal molecule were compared: random, for a baseline; greedy, where only the mean of the distribution was considered; and lower confidence bound (LCB, $\beta = 1$), which attempted to benefit from uncertainty estimates (Figure~\ref{fig:bo}, Appendix~\ref{app:bo}). As expected, greedy and LCB greatly surpassed the random acquisition function. In addition, greedy and LCB exhibited very similar performance, which suggests that the uncertainty estimates of CNPs did not offer a competitive advantage for molecular optimization. However, both methods found the best molecule in the dataset quickly, making it difficult to draw strong conclusions. A more challenging optimization task within a larger molecular library is left for future work.

\section{Discussion}

Our results demonstrate that CNPs have outstanding performance in few-shot learning of complex molecular properties such as docking scores. The application of CNPs to impute sparse chemical datasets could be highly impactful, even if one had confidence in only a small fraction of the imputations. However, the correct way to calibrate uncertainty estimates in CNPs is a question that requires further study, as is the potential to generalise to more diverse function classes. Applying similar models for probabilistic prediction, such as LNPs, in more complex imputation and molecular optimization tasks is an exciting area for future work.

\vfill\newpage
\bibliographystyle{unsrt}
\bibliography{references}

\setcounter{figure}{0}
\makeatletter 
\renewcommand{\thefigure}{A\@arabic\c@figure}
\makeatother

\setcounter{table}{0}
\makeatletter 
\renewcommand{\thetable}{A\@arabic\c@table}
\makeatother

\vfill\newpage
\begin{appendices}

\section{Probabilistic regression and calibration}
\label{app:regression_calibration}


\begin{figure}[h!]
\captionsetup[subfloat]{labelformat=empty}
\centering
\begin{minipage}{1\linewidth}
\begin{minipage}{.5\linewidth}
\centering
\subfloat[]{\label{fig:app_regression_calibration_a_improved}\includegraphics[scale=.5]{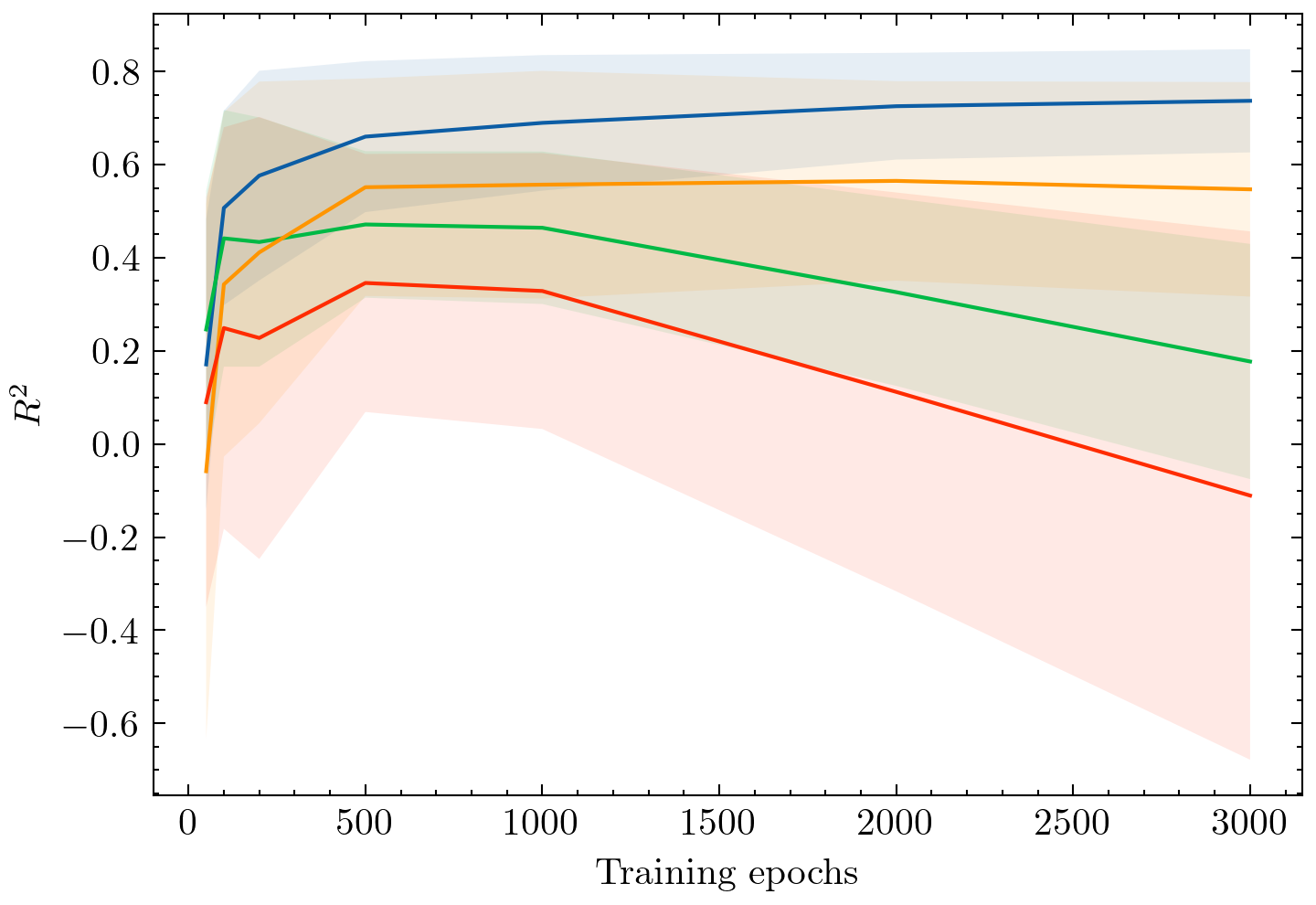}}
\end{minipage}%
\begin{minipage}{.5\linewidth}
\centering
\subfloat[]{\label{fig:app_regression_calibration_b_improved}\includegraphics[scale=.5]{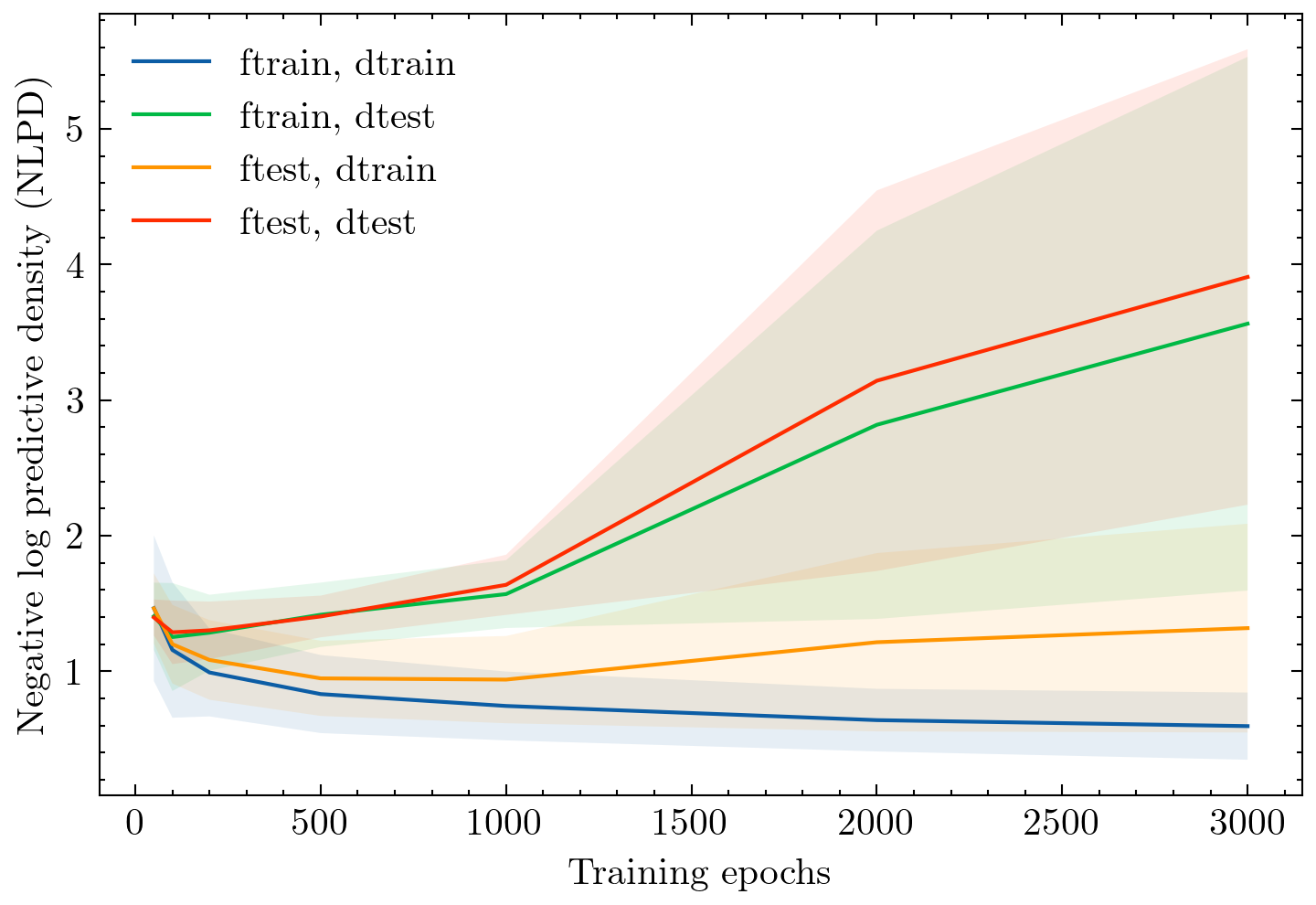}}
\end{minipage}
\end{minipage}
\caption{Performance of the CNP on the different data splits as the number of training epochs increases. The shaded regions indicate the standard deviation across different random initializations and protein targets in the same set \textit{ftrain}, \textit{ftest}, \textit{dtrain} or \textit{dtest}.}
\label{fig:app_regression_calibration_improved}
\end{figure}


\section{Low data}
\label{app:low_data}


\begin{figure}[h!]
\captionsetup[subfloat]{labelformat=empty}
\centering
\begin{minipage}{0.9\linewidth}
\begin{minipage}{.5\linewidth}
\centering
\subfloat[]{\label{fig:app_low_data_a}\includegraphics[scale=.5]{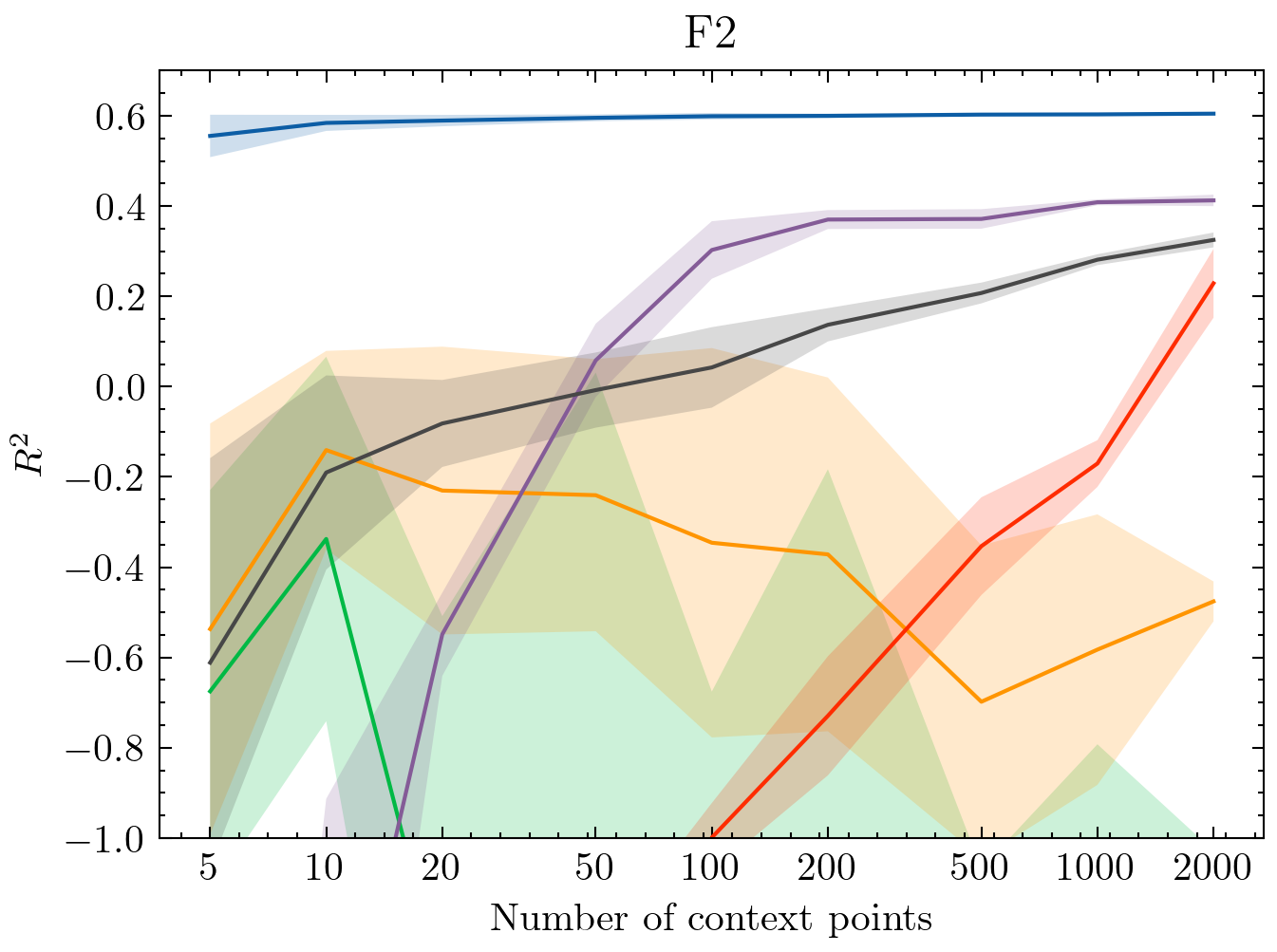}}
\end{minipage}
\begin{minipage}{.5\linewidth}
\centering
\subfloat[]{\label{fig:app_low_data_b}\includegraphics[scale=.5]{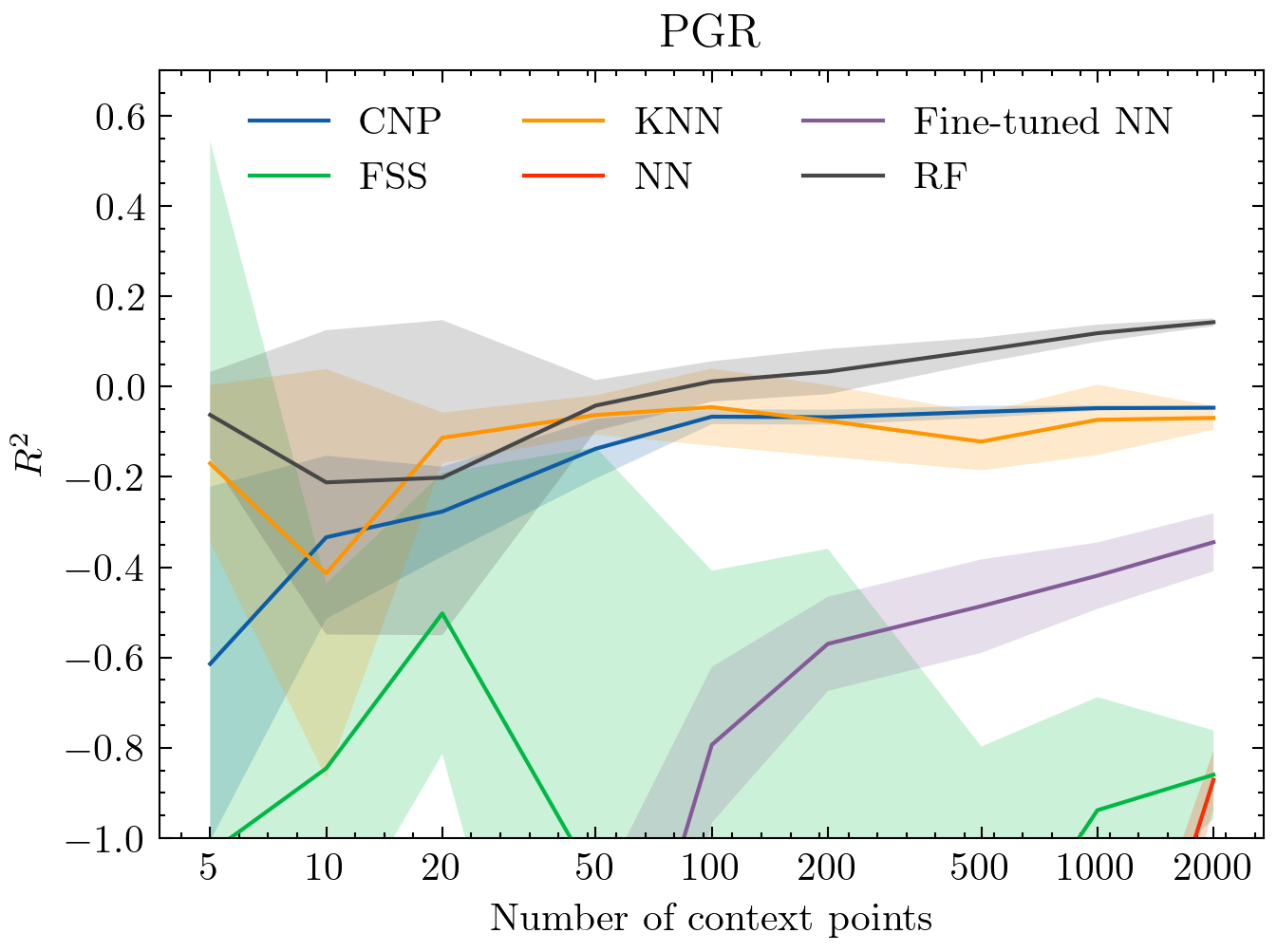}}
\end{minipage}
\end{minipage}
\begin{minipage}{0.9\linewidth}
\centering
\subfloat[]{\label{fig:app_low_data_c}\includegraphics[scale=.5]{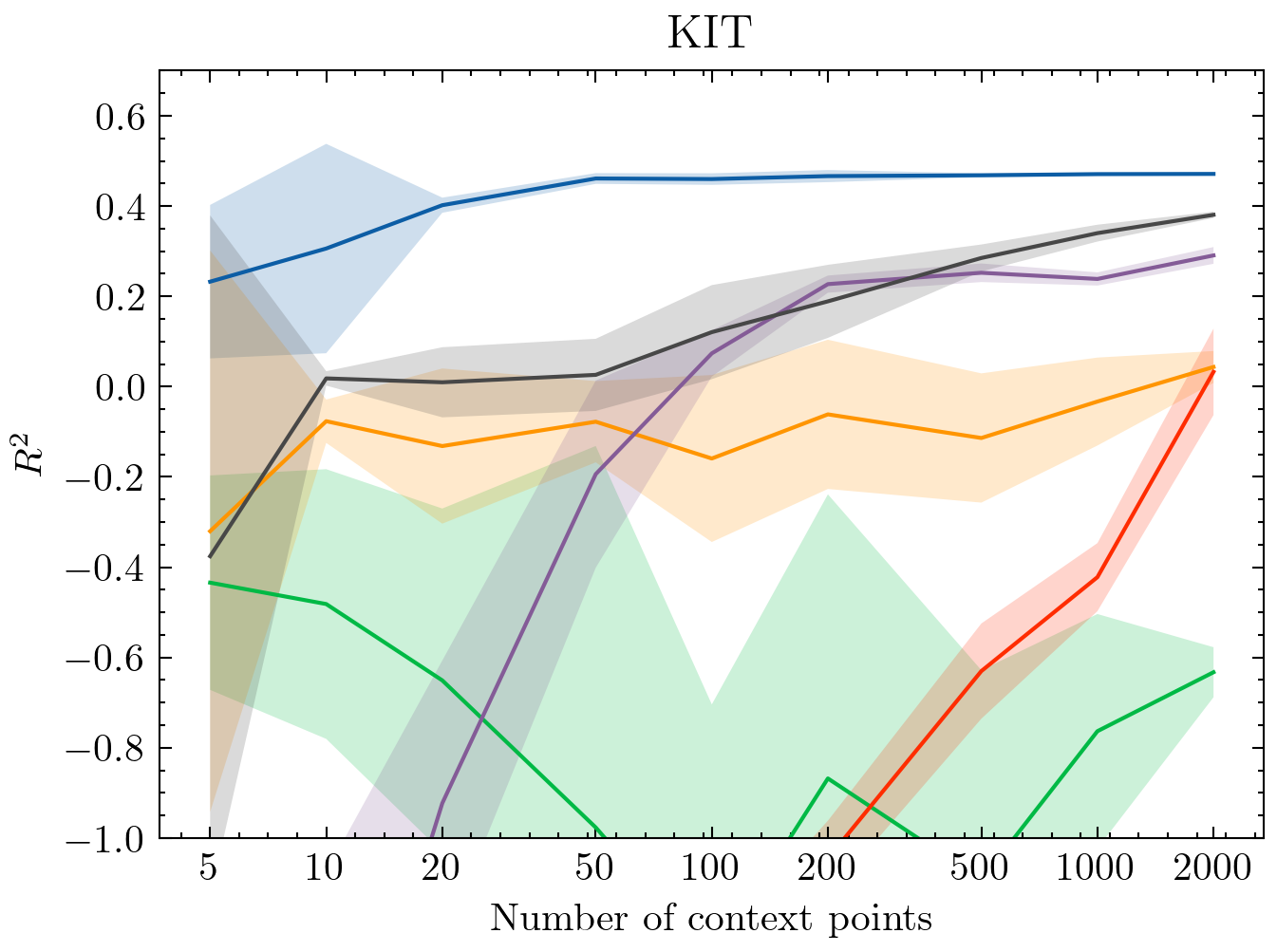}}
\end{minipage}
\caption{Few-shot performance of the CNP and benchmark models on three test functions in \textit{ftest}, ranging in difficulty from easy (F2), medium (KIT), and hard (PGR). The x-axis shows the number of datapoints in \textit{ftest, dtrain} used as context by the CNP or as training points by the benchmarks. Performance was evaluated on \textit{ftest, dtest}.}
\end{figure}

\section{QED-modified docking scores}
\label{app:qed_regression}

The quantitative estimate of drug-likeness (QED) of a molecule is a coefficient between 0 and 1 that attempts to quantify the molecule's similarity to approved drugs. In some experiments, we combine docking scores and QED values to create a new artificial score. QED-modified scores are expected to be more challenging to predict and to reflect drug-likeness. 

Given a molecule $m$ with docking score $s(m, t)$ for protein target $t$, we define its QED-modified score $s'$ as

$$
s'(m, t)  := s(m, t) + 10 (1 - \text{QED}(m)).
$$

\vfill\newpage
\section{Bayesian optimization with CNPs}
\label{app:bo}

\begin{figure}[h!]
\captionsetup[subfloat]{labelformat=empty}
\centering
\begin{minipage}{0.9\linewidth}
\begin{minipage}{.5\linewidth}
\centering
\subfloat[]{\includegraphics[scale=.7]{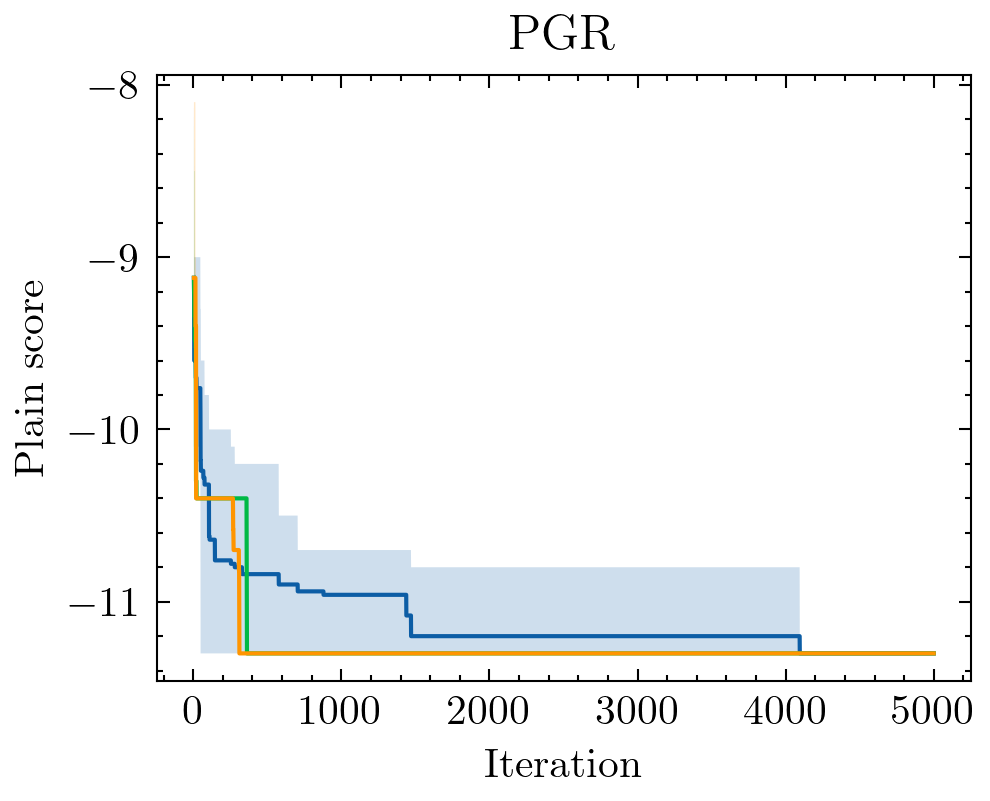}}
\end{minipage}%
\begin{minipage}{.5\linewidth}
\centering
\subfloat[]{\includegraphics[scale=.7]{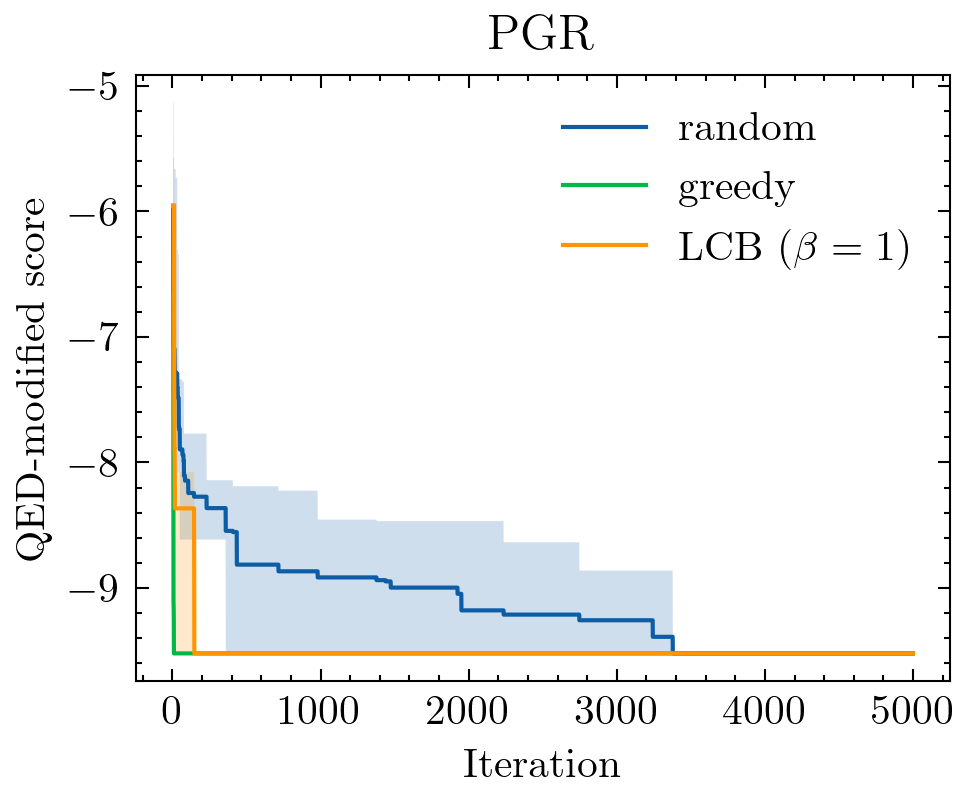}}
\end{minipage}
\begin{minipage}{.5\linewidth}
\centering
\subfloat[]{\includegraphics[scale=.7]{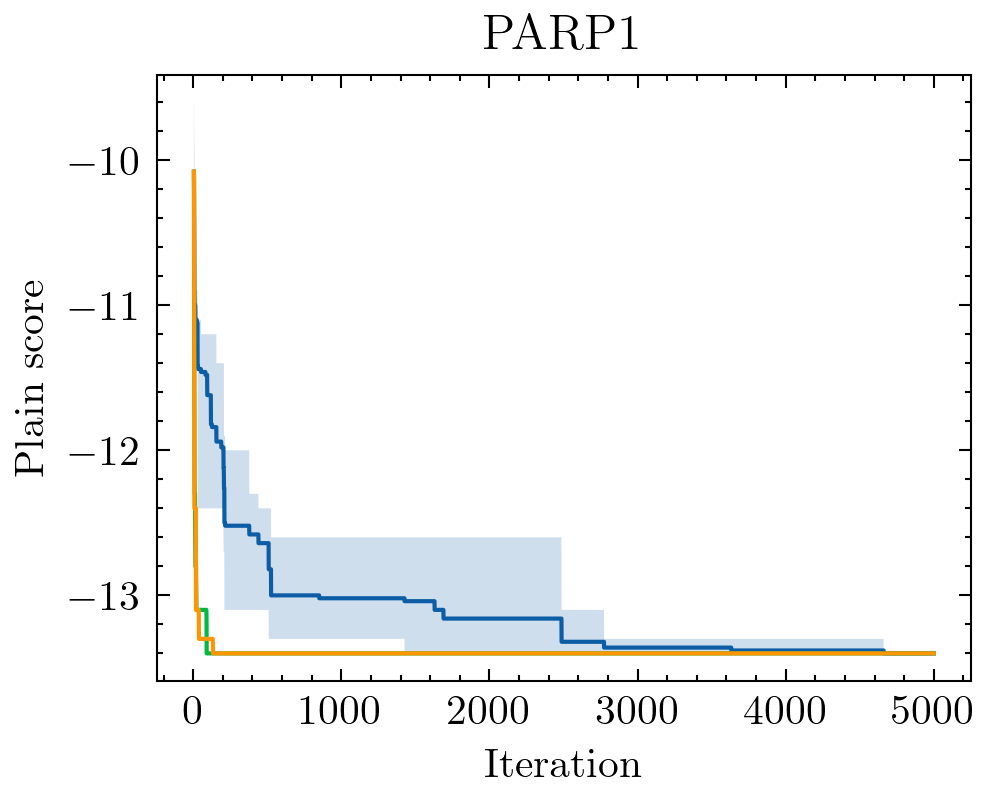}}
\end{minipage}
\begin{minipage}{.5\linewidth}
\centering
\subfloat[]{\includegraphics[scale=.7]{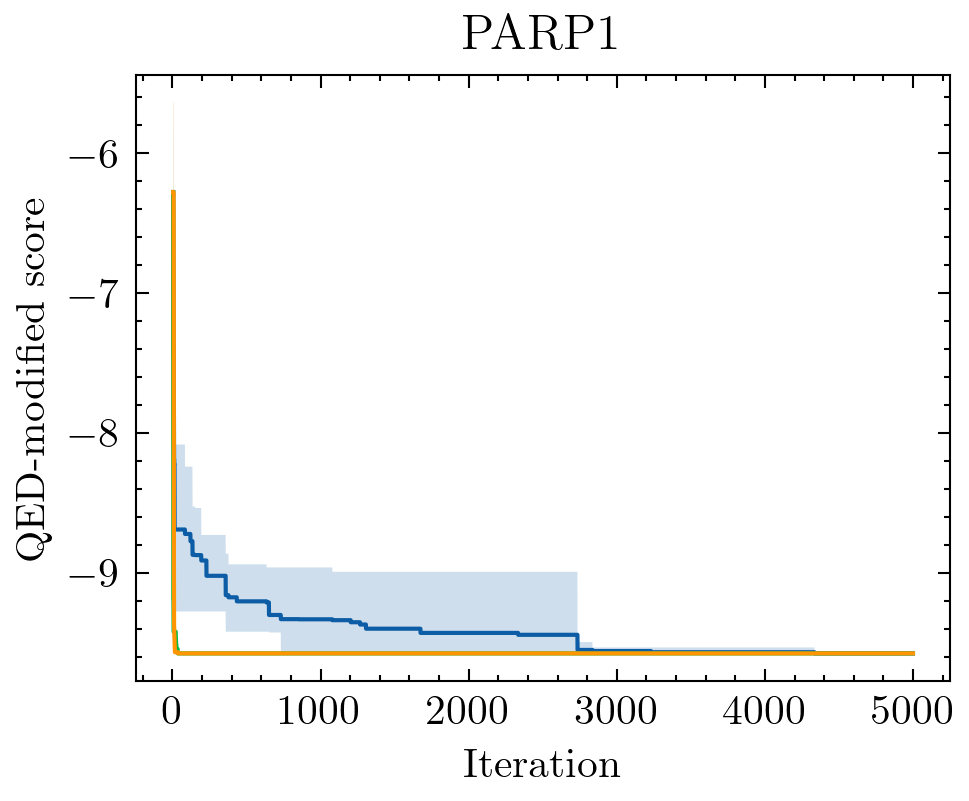}}
\end{minipage}
\begin{minipage}{.5\linewidth}
\centering
\subfloat[]{\includegraphics[scale=.7]{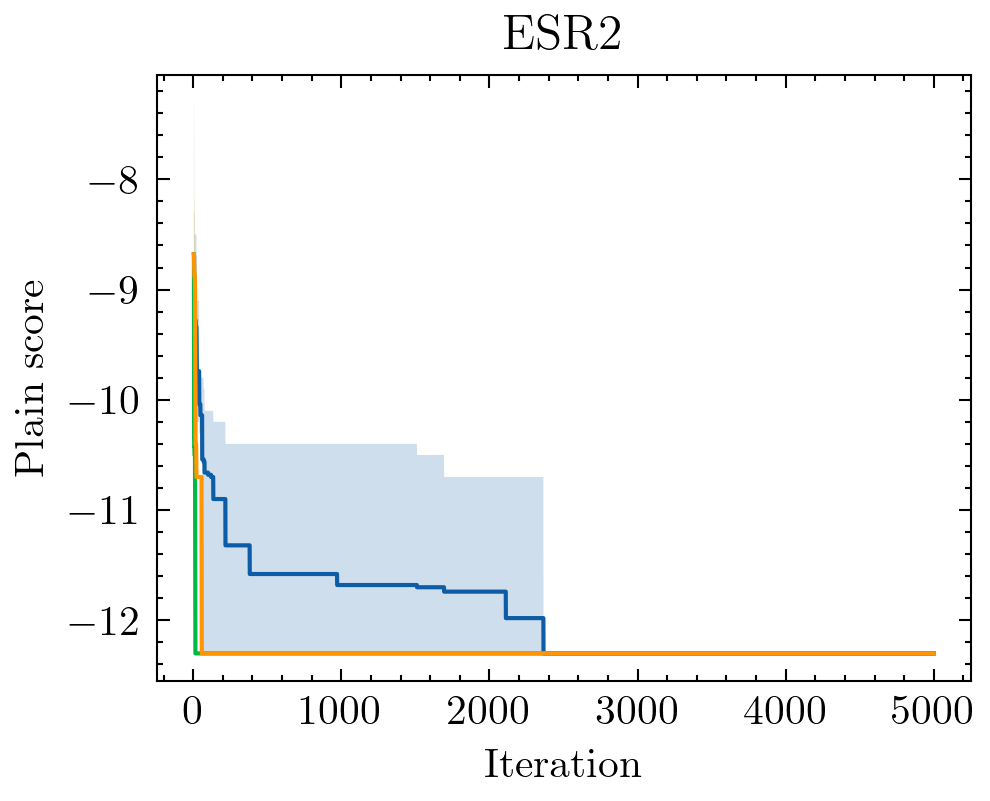}}
\end{minipage}
\begin{minipage}{.5\linewidth}
\centering
\subfloat[]{\includegraphics[scale=.7]{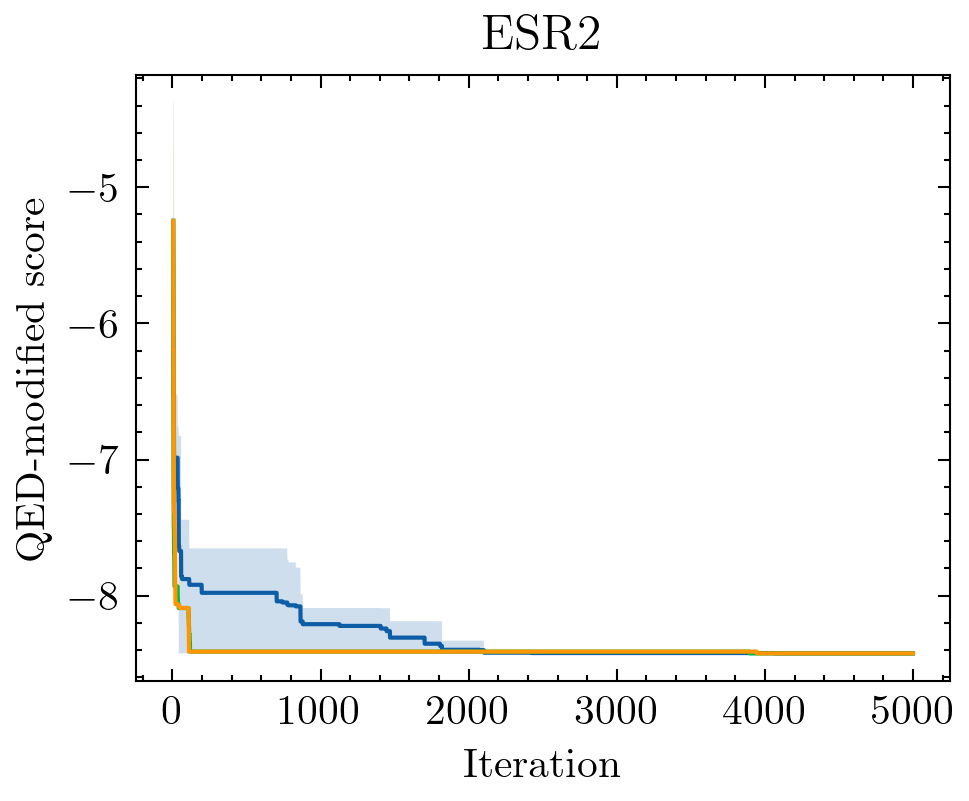}}
\end{minipage}
\begin{minipage}{.5\linewidth}
\centering
\subfloat[]{\includegraphics[scale=.7]{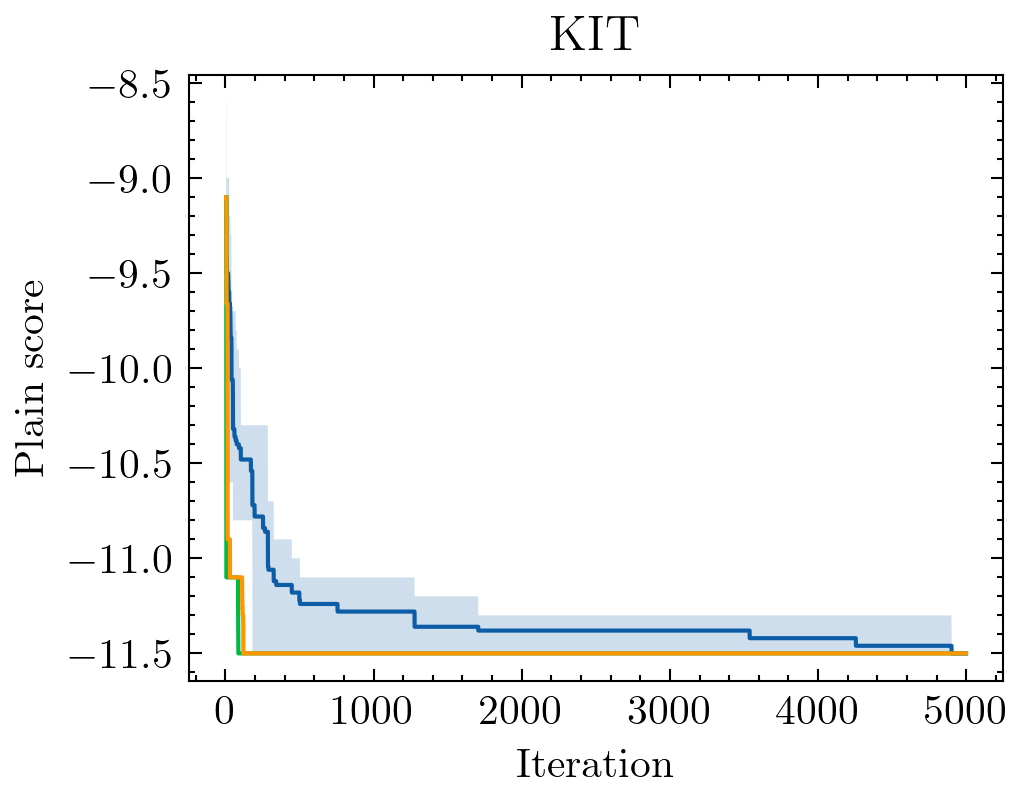}}
\end{minipage}
\begin{minipage}{.5\linewidth}
\centering
\subfloat[]{\includegraphics[scale=.7]{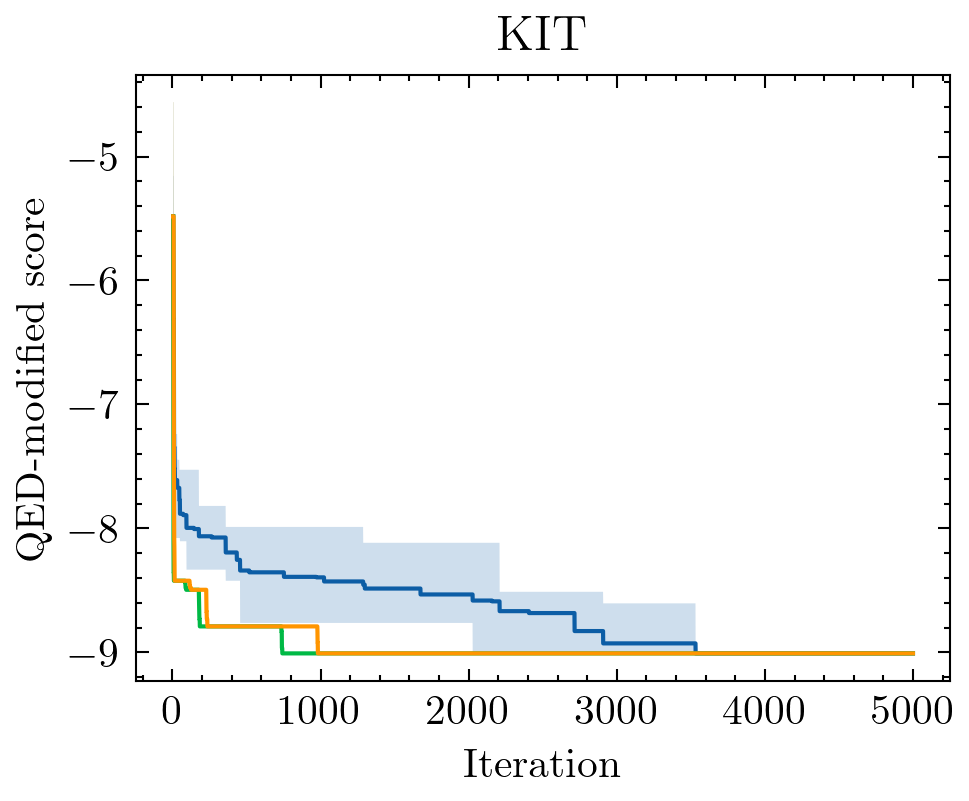}}
\end{minipage}
\end{minipage}
\label{fig:app_bo}
\end{figure}

\end{appendices}

\end{document}